\documentclass[sigconf]{acmart}
\makeatletter
\newif\if@restonecol
\makeatother

\usepackage[linesnumbered,ruled,vlined]{algorithm2e}
\usepackage{algpseudocode}
\usepackage{amsmath}
\usepackage{color}

\AtBeginDocument{%
  \providecommand\BibTeX{{%
    \normalfont B\kern-0.5em{\scshape i\kern-0.25em b}\kern-0.8em\TeX}}}

\setcopyright{acmcopyright}
\copyrightyear{2020}
\acmYear{2020}
\acmDOI{https://doi.org/10.1145/3394171.3413746}

\acmConference[MM ’20]{}{October 12–16, 2020}{Seattle, WA, USA}
\acmBooktitle{MM ’20: 
  October 12–16, 2020, Seattle, WA, USA}
\acmPrice{15.00}
\acmISBN{ACM ISBN 978-1-4503-7988-5/20/10}

\begin{document}

\title{Relational Graph Learning for \\ Grounded Video Description Generation}

\author{Wenqiao Zhang}
\affiliation{\institution{Zhejiang University}}
\email{wenqiaozhang@zju.edu.cn}

\author{Xin Eric Wang}
\affiliation{\institution{University of California,
Santa Cruz}}
\email{xwang366@ucsc.edu}

\author{Siliang Tang}
\authornote{Siliang Tang is the corresponding author.}
\affiliation{\institution{Zhejiang Universety}}
\email{siliang@zju.edu.cn}

\author{Haizhou Shi, Haocheng Shi}
\affiliation{\institution{Zhejiang University}}
\email{shihaizhou,hcshi@zju.edu.cn}

\author{Jun Xiao,Yueting Zhuang}
\affiliation{\institution{Zhejiang University}}
\email{junx,yzhuang@cs.zju.edu.cn}

\author{William Yang Wang}
\affiliation{\institution{University of California,
Santa Barbara}}
\email{william@cs.ucsb.edu}

\renewcommand{\shortauthors}{Trovato and Tobin, et al.}

\begin{abstract}
Grounded video description (GVD) encourages captioning models to attend to appropriate video regions (\emph{e.g., objects}) dynamically and generate a description. Such a setting can help explain the decisions of captioning models and prevents the model from hallucinating object words in its description. However, such design mainly focuses on object word generation and thus may ignore fine-grained information and suffer from missing visual concepts. Moreover, relational words (\emph{e.g., ``jump left or right''}) are usual spatio-temporal inference results, \emph{i.e.}, these words cannot be grounded on certain spatial regions. To tackle the above limitations, we design a novel relational graph learning framework for GVD, in which a language-refined scene graph representation is designed to explore fine-grained visual concepts. Furthermore, the refined graph can be regarded as relational inductive knowledge to assist captioning models in selecting the relevant information it needs to generate correct words. We validate the effectiveness of our model through automatic metrics and human evaluation, and the results indicate that our approach can generate more fine-grained and accurate description, and it solves the problem of object hallucination to some extent.
\end{abstract}

\ccsdesc[500]{Computing methodologies}
\ccsdesc[500]{Computing methodologies~Artificial intelligence}
\ccsdesc[500]{Computing methodologies~Scene understanding}
\ccsdesc[500]{Computing methodologies~Computer vision tasks}
\ccsdesc[500]{Computing methodologies~Natural language generation}

\keywords{grounded video description, language refined scene graph, object hallucination, fine-grained information}


\maketitle

\section{Introduction}
Recently, video captioning~\cite{kojima2002natural}, the task of automatically generating a sequence of natural-language words to describe a video, has drawn increasing attention~\cite{mun2019streamlined,park2019adversarial, wang2018video}. However, these models are known to have poor grounding performance, which leads to \emph{objects hallucination}~\cite{rohrbach2018object}. Captioning models generate descriptive objects that are not in the video because of similar semantic contexts or pre-extracted priors during the training stage. As a result, some models lack interpretability even if they have high captioning scores. Therefore, grounded video description (GVD)~\cite{zhou2019grounded}, which tries to improve the grounding performance of captioning models, has been proposed. In this approach, the captioning model learns to ground related video regions (\emph{e.g., objects}) that are used as input to predict the next word. Such a setting can teach models to rely explicitly on the corresponding evidence in the video when generating descriptions.

Focusing on object word generation can only partially address the problem of object hallucination. Furthermore, this design may lead to some potential limitations. First, a GVD model encourages captioning models to focus on related regions of objects to generate correct words; thus, the fine details (\emph{e.g., related objects and attributes of grounded object}) in the corresponding video may be ignored, and may generate a coarse-grained descriptive sentence. Second, relational words (relationships among or between salient objects), such as ``\emph{climb up or down}'', are usual inference results from sequential frames. No specific spatial regions can ground these words, thus, using only grounding regions may lead to inappropriate word generation. Therefore, an optimization goal of GVD is to produce fine-grained information and a reasonable learning approach to describing a video with fine and correct details.

\begin{figure}[t]
\includegraphics[width=0.5\textwidth]{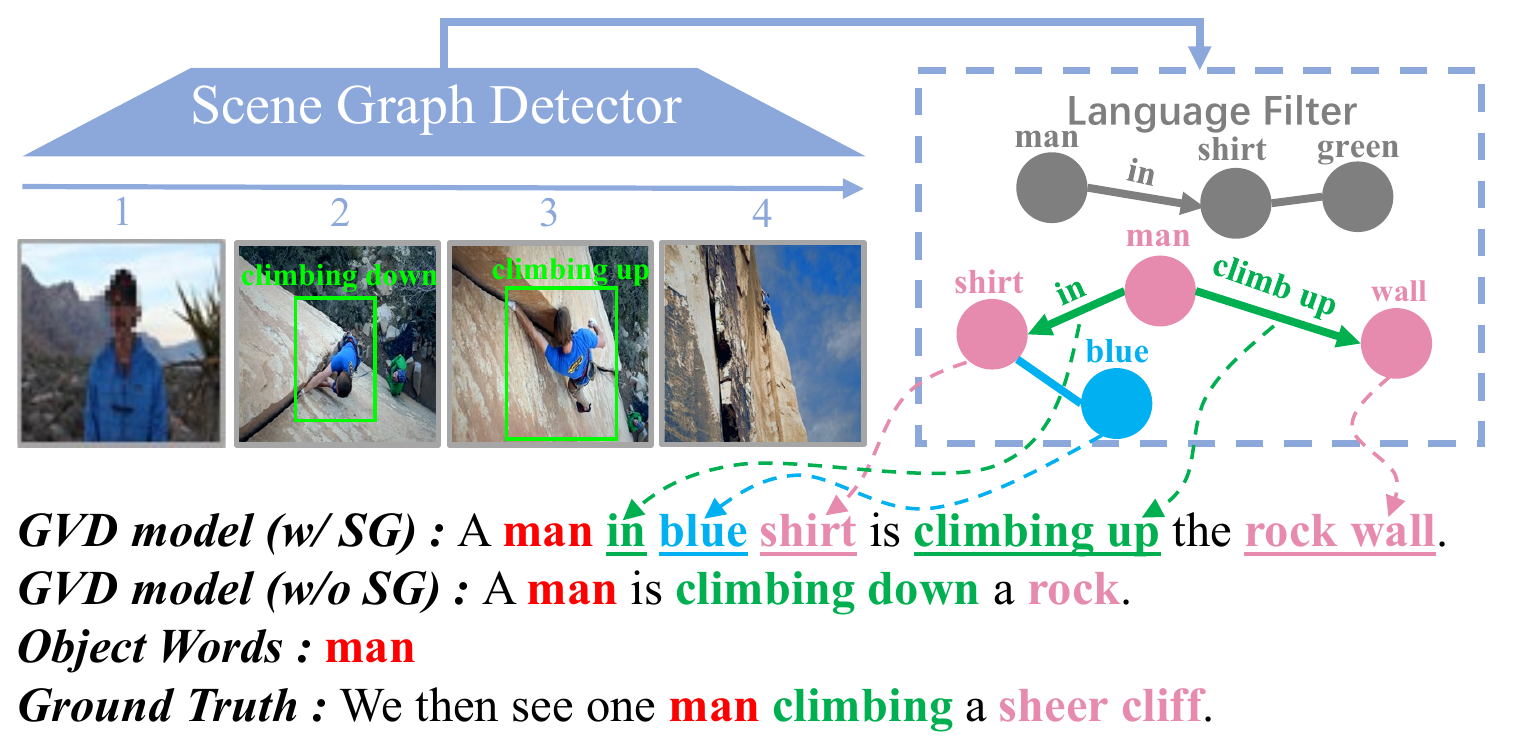}
\centering\caption{An example of how the language-refined graph facilitates GVD generation. The green bounding box is the attended region. Pink, green and blue words represent the objects, relationships and attributes.}
\end{figure}

Inspired by recent breakthroughs in higher-level visual understanding tasks, scene graph ($\mathcal{SG}$ ) construction has become a popular area with remarkable advancements.~\cite{ding2019boundary,ding2019semantic}. As an abstraction of objects and their complex relationships, $\mathcal{SG}$ can provide rich semantic information of the visual domain. Thus, $\mathcal{SG}$  is expected to deal with the problems mentioned above.  On the one hand, $\mathcal{SG}$ provides complementary information to assist the captioning model in generating fine-grained phrases, such as ``\emph{man in blue shirt}'' (Figure 1). On the other hand, $\mathcal{SG}$ with relational inductive knowledge guides the captioning model to generate appropriate relational words, such as ``\emph{climb up}'', and ground the correct region in the third frame. While  w/o $\mathcal{SG}$ grounding a similar region in the second frame but generate inappropriate relational words  ``\emph{climb down}''. However, directly using $\mathcal{SG}$ does not solve the problem of semantic inconsistency between predefined concept categories and target lexical words. For example, in Figure~1, $\langle$ \emph{man-in-shirt} $\rangle$ and $\langle$ \emph{shirt-green} $\rangle$ are the unconcerned concepts.We notice that a sentence (ground truth of video description) can also parse a language $\mathcal{SG}$~\cite{anderson2016spice} that contains the key visual concepts. The language $\mathcal{SG}$ can be treated as linguistic guidance to refine the visual $\mathcal{SG}$. Exactly as shown in Figure 1, the unconcerned concepts with color gray are filtered and do not hallucinate objects in the description.h

Based on this insight, we propose the relational graph learning framework (RGL), which incorporates $\mathcal{SG}$ into the conventional encoder-decoder method for GVD generation. Specifically, given a video, the training pipeline has three parts: 1) \emph{Relational Graph Encoder}. We first build the frame scene graph $\mathcal{SG^{F}}$ and the language scene graph $\mathcal{SG^{L}}$ from the video frames and its ground truth at the training stage.  Then, to filter the unconcerned concepts, we regard the  $\mathcal{SG^{L}}$ as linguistic guidance to refine the $\mathcal{SG^{F}}$. Thus, the refined graph representation $\mathcal{SG^{R}}$ with relational inductive knowledge is obtained for improved GVD generation. 2) \emph{Sentence Decoder}. We introduce a selection mechanism for a sentence generator that learns to decide the utilization of grounding regions and refined graph $\mathcal{SG^{R}}$. Such a decoder can generate video descriptions reasonably on a fine-grained level. In addition, we derive a context generator with refined graph $\mathcal{SG^{R}}$ to leverage and update the compressed semantic information and maintain coherence among sentences.  3)  \emph{Grounding Module}. We build a grounding and localizing mechanism, which not only encourages the model to ground the regions dynamically on the basis of the current semantic context to predict words but also localizes regions using the generated object words. Such a setting can boost the accuracy of the object word generation and address the problem of object hallucination. In summary, the major contributions of our study are as follows:

 \begin{itemize}
\item We develop a language-refined $\mathcal{SG}$ representation that contains the key visual concepts for GVD generation.

\item We propose a captioning model with a selection mechanism that selects the relevant information it needs on the basis of the current semantic context to describe a video reasonably.

\item We demonstrate the superiority of RGL in generating fine-grained and accurate description via automatic metrics and human evaluation.

\item This attempt is the first to combine GVD and visual $\mathcal{SG}$ organically. The proposed RGL can improve captioning quality, grounding performance and alleviate object hallucination simultaneously.
 \end{itemize}

\section{Related Work}

\noindent$\textbf{Video Captioning}$: Video captioning is being actively studied in vision and language research. The prevailing video captioning techniques often incorporate the encoder-decoder pipeline inspired by the first successful sequence-to-sequence model S2VT~\cite{venugopalan2015sequence}. Benefitting from the rapid development of deep learning, video captioning models have achieved remarkable advances using attention mechanism~\cite{yao2015describing,song2017hierarchical,yan2019stat,zhang2021tell}, memory networks~\cite{wang2018m3,chen2017video,li2018multimodal,pei2019memory}, reinforcement learning~\cite{wang2018video,li2019end,pasunuru2017reinforced} and generative adversarial networks~\cite{park2019adversarial,yang2018video}. Although these encoder-decoder-based methods have reached impressive performance on automatic metrics, they often neglect how well the generated caption words (e.g., objects) are grounded in the video, making models less explainable and trustworthy.

\noindent$\textbf{Visual Grounding}$: Visual grounding models encourage captioning generators to link phrases with specific spatial regions of images or videos, thereby presenting a potential way to improve the explainability of models~\cite{rohrbach2016grounding,fukui2016multimodal,xiao2017weakly,zanfir2016spatio,zhou2019grounded,zhang2021consensus}.  The most common way of grounding models is to predict the next word using an attention mechanism, which is deployed over noun phrases, with supervised bounding boxes as input. However, these models often focus on object word generation and further produce a coarse-grained sentence to describe an image or a video. Moreover, grounding video regions seem inappropriate to predict relational words because these words do not usually correspond to a specific spatial region.

\begin{figure*}[t]
\includegraphics[width=1\textwidth]{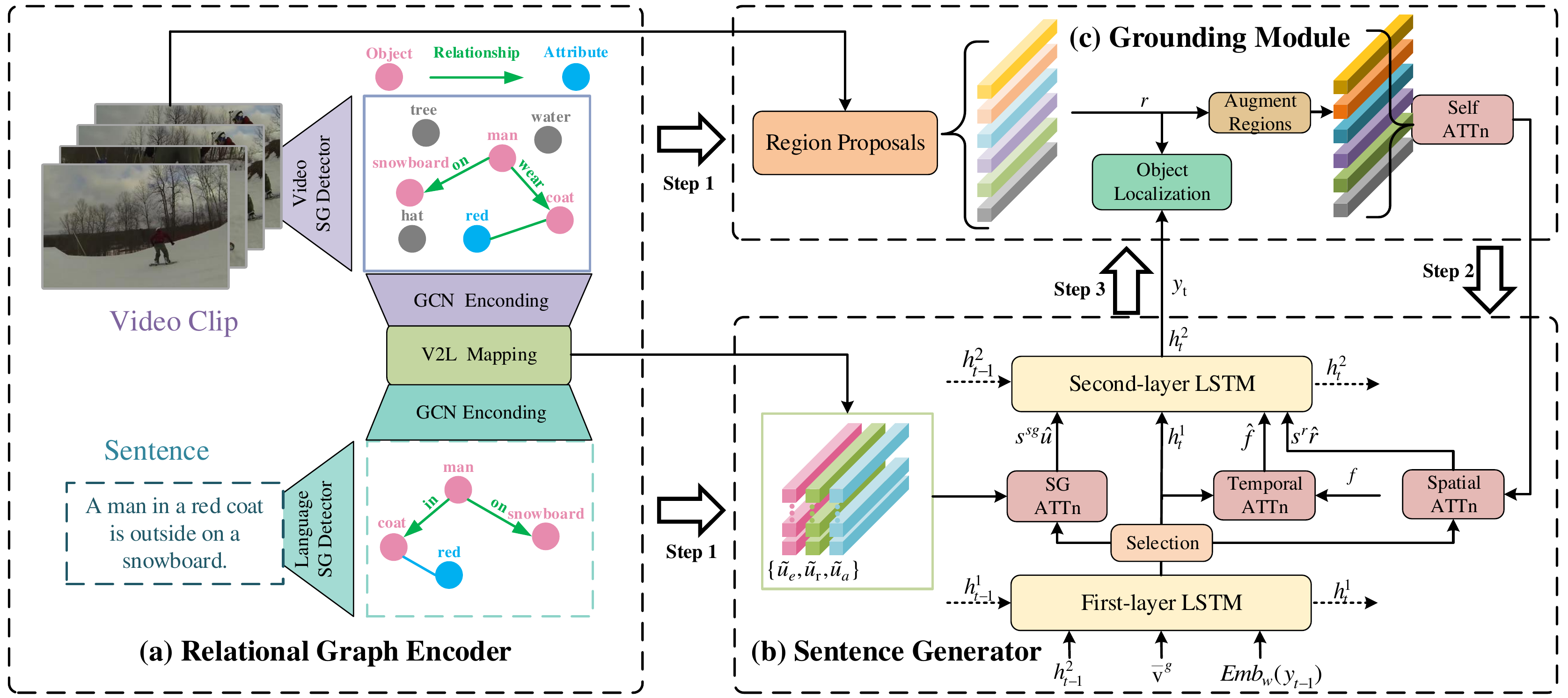}
\centering\caption{Overview of our method. It consists of three modules: (a) The relational graph encoder to produce the language-refined graph representation $\tilde{\textbf{u}}$ from the source video and its ground truth. (b) The sentence generator uses a stacked two-layer LSTM with a selection mechanism for video description generation. (3) The grounding module has the mechanism of region grounding and object localization to predict the correct object words. }
\end{figure*}

\noindent$\textbf{Scene Graph}$: Recently, $\mathcal{SG}$ construction have become popular research topics with significant advancements  ~\cite{zhuang2017towards,zellers2018neural,yang2018graph,gu2019scene,xu2017scene,zhang2017visual} based on the Visual Genome~\cite{krishna2017visual} dataset.  The $\mathcal{SG}$s contain structured semantic information and can represent scenes as directed graphs, where nodes are objects and edges are relationships and attributes. Using this inductive information is natural to improve the performance of vision-language tasks, e.g., image captioning~\cite{yang2019auto,gu2019unpaired}, VQA~\cite{shi2019explainable,teney2017graph}. However, directly feeding the $\mathcal{SG}$ to the captioning model may lead to a noncorrespondence problem between vision and language. Therefore, how to refine the relational knowledge is key to promote the vision-language field further.

\noindent$\textbf{Graph Convolutional Network}$: A graph convolutional network (GCN)~\cite{berg2017graph}is a weighted average function that operates directly on a graph and induces the embedding vectors of nodes on the basis of the properties of their neighborhoods. Graph neural networks have achieved remarkable success in processing graph-structured data and have been widely adopted in many areas, such as semantic role labeling~\cite{marcheggiani2017encoding,marcheggiani2018exploiting}, relation classification~\cite{zhou2016attention,wang2016relation}, social network mining~\cite{derr2018signed}, text classification~\cite{yao2019graph,yao2019clinical}, recommendation system~\cite{ying2018graph}, and scene understanding~\cite{yang2018graph}.  Given the effectiveness of GCN in handling graph-based data, we use a GCN to capture the contextual relations among the $\mathcal{SG}$s of a video for GVD generation.

\section{Method}
\subsection{Task Description}
Before presenting our method, we first introduce some basic notions and terminologies. Given a video \textbf{\emph{V}}=$\{ \textbf{\emph{V}}_1, \cdots, \textbf{\emph{V}}_L$\} consisting of \emph{L} clips, the goal of grounded video description (GVD) is to generate the natural language description \textbf{\emph{S}}=$\{ \textbf{\emph{S}}_1, \cdots, \textbf{\emph{S}}_L$\}, and localize the object words \textbf{\emph{O}}=$\{ \textbf{\emph{O}}_1,$ $\cdots, \textbf{\emph{O}}_K$\} in frames, where \emph{K} is the number of objects appearing in the description. We denote the model as \emph{G}, and the sentence $\hat{\textbf{\emph{S}}}$, object grounding $\hat{\textbf{\emph{O}}^g}$  generated by  \emph{G(\textbf{V})}, \emph{i.e.}, $\hat{\textbf{\emph{S}}}$, $\hat{\textbf{\emph{O}}^g}$ = \emph{G(\textbf{V})}.  For a model parameterized by $\theta$, we define the loss for a training pair as $\mathcal{L}$(($\textbf{\emph{S}}$, $\textbf{\emph{O}}^g$), $G$(\textbf{\emph{ V}}; $\Theta$)).

Figure~2 shows the overall pipeline of our Relational Graph Learning (RGL) approach, which mainly consists of the \emph{Relational Graph Encoder}, \emph{Sentence Decoder} and \emph{Grounding Module}. (1) Given a video and its corresponding ground truth, the \emph{Relational Graph Encoder} module first transforms them into the $\mathcal{SG}$s. Then adapting the visual-language mapping is adopted to generate a language-refined graph representation $\tilde{\textbf{u}}$ (Section 3.2). (2) The \emph{Sentence Decoder} is encouraged to select information dynamically from $\tilde{\textbf{u}}$ and ground regions $\textbf{r}$ to generate the current word. (Section 3.3). (3) We compute the localization accuracy of object words on the ground truth sentence, and the language model dynamically attends region proposals $\textbf{r}$ in the subsequent word prediction stage. (Section 3.4).For clarity, we use a GVD task as an example to illustrate our method.

\subsection{Relational Graph Encoder}
Generally, the $\mathcal{SG}$ defined in our task $\mathcal{SG} = (\mathcal{N}, \mathcal{E})$ and contains a
set of nodes $\mathcal{N}$ and  edges $\mathcal{E}$. As exemplified in Figure~2(a), the node set $\mathcal{N}$ contains three types of nodes: object node $o$, attribute
node $a$, and relationship node $r$.

\subsubsection{Scene Graph Detector}

Given a video clip, we sample several frames  to generate the \emph{Frame Scene Graph} $\mathcal{SG^F}$.  In detail, for $i^{th}$  frame,  the corresponding  $\mathcal{SG}_{i}^\mathcal{F}$ is extracted by the $\mathcal{SG}$ parser comprising an object detector Faster-RCNN~\cite{wang2017fast}, an attribute classifier~\cite{yang2019auto}, and a relationship classifier MOTIFS~\cite{zellers2018neural}. Hence, $\mathcal{SG^F} = \{\mathcal{SG}_{1}^\mathcal{F},\cdots, \mathcal{SG}_{Q}^\mathcal{F} \}$, where $Q$ is number of sampled frames. To generate the \emph{Language Scene Graph} $\mathcal{SG^L}$, we use the $\mathcal{SG}$ generator~\cite{anderson2016spice} that parses the sentence to a syntactic dependency tree. Then, the rule-based method~\cite{schuster2015generating} is developed to transform the tree into an $\mathcal{SG}$.

\subsubsection{Scene Graph Encoder}
To represent the nodes of the $\mathcal{SG}$, we denote the nodes $\{o, a, r\}$ by the label embeddings as $\{\textbf{e}^o, \textbf{e}^a, \textbf{e}^r\}$ $\in \mathbb{R}^{e}$, corresponding respectively to objects, attributes and relationships both in  $\mathcal{SG^F}$ and $\mathcal{SG^L}$ . In particular, different from using the simple label embeddings $ \textbf{e}^o$ and $\textbf{e}^r$ in $\mathcal{SG^F}$, $\textbf{e}^o$ is the key to connect domains of vision and language. Thus, we introduce the Multi-modal Factorized Bilinear Pooling (MFB)~\cite{yu2017multi} to fuse the region features and label embeddings to augment object representation $\textbf{e}^o$, which is known to be effective in multi-modal tasks~\cite{liu2018towards,li2018essay}.

\noindent$\textbf{Node embedding}:$ To encode the $\mathcal{SG}$ nodes at a unified representation $\textbf{u}=\{\textbf{u}^o, \textbf{u}^a, \textbf{u}^r\}$ $\in \mathbb{R}^{u}$, we introduce the Graph Convolutional Network (\emph{GCN})~\cite{marcheggiani2017encoding}, which can embed the graph structure into vector representations. Thus, we use the GCN that encodes three kinds of node embeddings by considering their neighborhood information. All the \emph{GCNs} are defined with the same structure but independent parameters.

\noindent$\textbf{Objects Encoding}:$ An object $o_j$ in $\mathcal{SG}$, it can play different roles (``\emph{subject}'' or ``\emph{object}'') due to different edge directions, \emph{i.e.}, two triples with different relationships, $\langle {o_i} - r_{i,j} - {o_j} \rangle$ and $\langle o_j - r_{j,k} - o_k \rangle$. Such associated objects by the cascaded encoding scheme can represent the global information of objects for a frame or sentence. Therefore we compute $\textbf{u}^{o_j}$ via explicit role modeling:
\begin{equation}\label{2}
\begin{aligned}
\textbf{u}^{o_j} =& \frac{1}{N_{o_j}} [\sum_{\textbf{e}^{o_j} \in sub} G_s(\textbf{e}^{o_j}, \textbf{e}^{r_{j,k}} , \textbf{e}^{o_k} )
\\ & + \sum_{\textbf{e}^{o_j} \in obj } G_o(\textbf{e}^{o_i}, \textbf{e}^{r_{i,j}}, \textbf{e}^{o_j} ) ]
\end{aligned}
\end{equation}
where $N_{o_j}$ = $ \left|sub(o_j) \right| +\left|obj(o_j) \right|$ is the total number of the relationship triplets in $\mathcal{SG}$  that object $o_j$ has. $G_s$ and $G_o$ are the convolutional operation for objects as a ``\emph{subject}'' or an ``\emph{object}''.

\noindent$\textbf{Attributes Encoding}:$ An $o_i$ in $\mathcal{SG}$  usually includes several attributes \{$a_1^i, \cdots, a^i_{T_{a_i}} $\}, where $T_{a_i}$ is the total number of attributes. Therefore, the unified $\textbf{u}^{a_i}$ can be computed as:
\begin{equation}\label{3}
\begin{aligned}
\textbf{u}^{a_i} =& \frac{1}{T_{a_i}} \sum_{i \in T_{a_i}} G_a(\textbf{e}^{o_i}, \textbf{e}^{a_{i}})
\end{aligned}
\end{equation}
where $G_a$ is the convolutional operation for object $o_i$ and its attributes.

\noindent$\textbf{Relationships Encoding}:$ The relationship between two salient objects $o_i$ and $o_j$ is given by the triplet $\langle o_i - r_{i,j} - o_j\rangle$. Similarly, the unified relationship encoding $\textbf{u}^{r_{i,j}}$ is produced as follows:
\begin{equation}\label{4}
\begin{aligned}
\textbf{u}^{r_{i,j}} =&  G^r_{i,j}(\textbf{e}^{o_i}, \textbf{e}^{r_{i,j}}, \textbf{e}^{o_j} )
\end{aligned}
\end{equation}
where $G^r_{i,j}$ is the convolutional operation for relational object $o_i$ and $o_j$.
\subsubsection{Visual-Language Mapping}
To refine the visual $\mathcal{SG^F}$, we adapt the visual domain to the language domain using the across-modality mapping function. In our model, we translate $\textbf{u}^\mathcal{F}$ (video) on the basis of  linguistic guidance $\textbf{u}^\mathcal{L}$ (sentence) to a refined  $\mathcal{SG}$ representation $\tilde{\textbf{u}}=\{\tilde{\textbf{u}}^o, \tilde{\textbf{u}}^a, \tilde{\textbf{u}}^r\}$ using an effective multi-layer perceptron. The visual-language mapping  loss $\mathcal{L}(M)$ is computed by mean squared error (MSE) loss as follows:
\begin{equation}\label{5}
\begin{aligned}
\mathcal{L}(M)= MSE(\textbf{u}^\mathcal{F}, \textbf{u}^\mathcal{L})
\end{aligned}
\end{equation}

 Such language-refined $\mathcal{SG}$ $\tilde{\textbf{u}}$ is then fed into the sentence decoder, providing relational inductive knowledge to generate GVD.

\subsection{Sentence Decoder}
\subsubsection{Single Sentence Generator}

We develop a stacked two-layer LSTM network using a selection mechanism to generate video description. The first layer of the LSTM network is for encoding the video features.  Specifically, we concatenate the previous hidden state $\textbf{h}_{t-1}^{2}$, global video representation $\bar{\textbf{v}}^{g}$, and embed the previously generated word $\textbf{y}_{t-1}$ as the source input to the first-layer LSTM:
\begin{equation}
\textbf{h}_{t}^{1} ={\rm LSTM_{1st}} (\textbf{h}_{t-1}^{1}, [\textbf{h}_{t-1}^{2}, \bar{\textbf{v}}^{g}, Emb_w(\textbf{y}_{t-1})])
\end{equation}
where $[\cdot ,\cdot]$ indicates the row-wise concatenation. $Emb_w$ is the word embedding matrix. $\bar{\textbf{v}}^{g}$  contains the pooling result of $\textbf{v}^{g}$ and the positional information of current video clip.

The second layer of the  LSTM network is for sentence generation. To improve the robustness of the sentence generator,   we enrich the language input as four parts instead of simply utilizing $\textbf{h}_{t}^{1}$  as the source input. We first extract the frame features $\textbf{f}=\{\textbf{f}_1, \cdots, \textbf{f}_Q\}$ and design a temporal attention~\cite{yan2019stat,zhang2019frame,zhang2020photo} to determine the weight for each frame.  The attended frame representation is given by $\hat{\textbf{f}}$.

We augment  region proposals $\textbf{r}=\{\textbf{r}_1, \cdots, \textbf{r}_N\}$ using its positional encoding and region-class similarity matrix. To  model the relations between regions further, we deploy a self-attention~\cite{vaswani2017attention} layer for $\textbf{r}$, thereby  allowing information to pass across all regions in the sample frames. Thus, the augmented region representation is given by $\tilde{\textbf{r}}$.

\begin{figure}[t]
\includegraphics[width=0.5\textwidth]{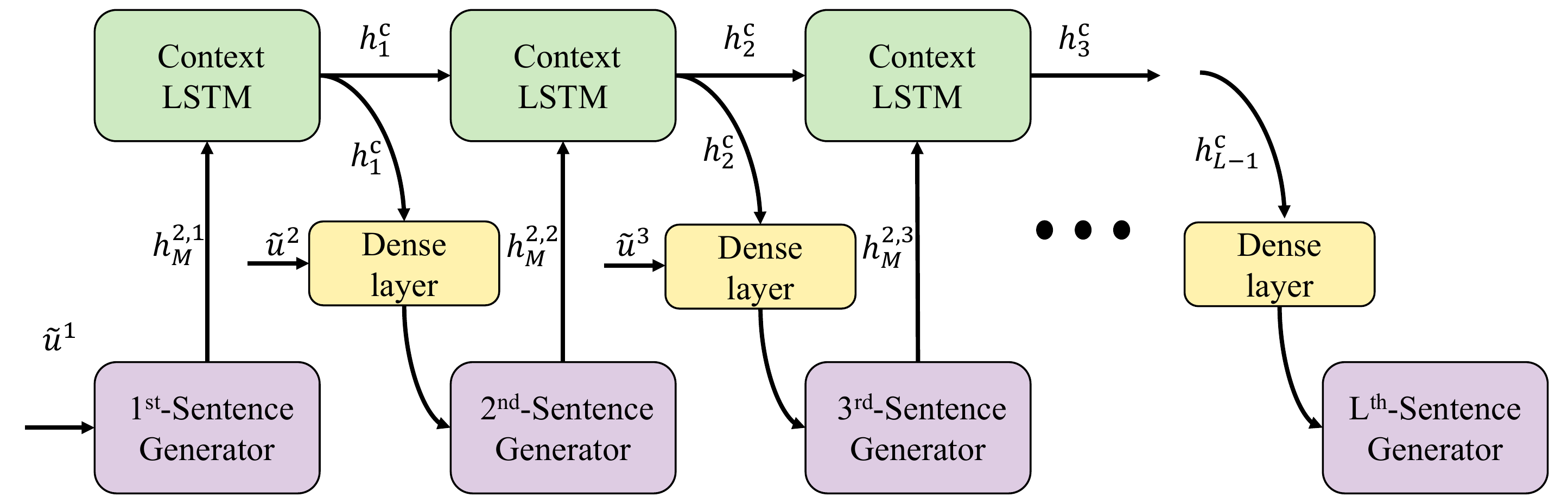}
\centering\caption{ Overview of the context generator.  The final hidden states in sentence generator are fed into context generator, whose output and $\mathcal{SG}$ representation $\tilde{\textbf{u}}$ are return to initialize the next sentence generator.}
\end{figure}

 \noindent$\textbf{Selection Mechanism}:$ The hidden state $\textbf{h}_{t}^{1}$ is determined to select dynamically \emph{how much } and \emph{what } from the  $\mathcal{SG}$ and grounding regions. (1) \emph{how much }, a soft attention is developed on $\tilde{\textbf{u}}$ and $\tilde{\textbf{r}}$,  and the corresponding weight score $s^r$ and $s^{sg}$ are determined to select dynamically \emph{how much} from $\mathcal{SG}$ and grounding regions. (2) \emph{what}, then we compute the spatial attention $a^r$ for $\{\tilde{\textbf{r}}_1, \cdots, \tilde{\textbf{r}}_N\}$ and soft attention $a^{sg}$ for $\{\tilde{\textbf{u}}^o, \tilde{\textbf{u}}^a, \tilde{\textbf{u}}^r\}$ to generate attended representation $\hat{\textbf{r}}$ and $\hat{\textbf{u}}$. $a^r$ and $a^{sg}$ are determined to select \emph{what } in $\tilde{\textbf{u}}$ and $\tilde{\textbf{r}}$. Lastly, we concatenate  $\textbf{h}_{t-1}^{1}$ ,$\hat{\textbf{f}}$,  $s^{sg} \hat{\textbf{u}}$ and  $s^{r} \tilde{\textbf{r}}$ as the input to the second layer of the LSTM network:
\begin{equation}
\textbf{h}_{t}^{2} = {\rm LSTM_{2nd}} (\textbf{h}_{t-1}^{2},[\textbf{h}_{t}^{1},\hat{\textbf{f}}, s^{sg} \hat{\textbf{u}}, s^r \hat{\textbf{r}}])
\end{equation}

Using the notation $\textbf{y}_{1:T}$, we refer to a sequence of words $\{\textbf{y}_1, \cdots, \textbf{y}_T \}$. For each step $t$, the conditional distribution over possible words is given by:
\begin{equation}\label{8}
p(\textbf{y}_t | \textbf{y}_{1:t-1}, \textbf{r}, \textbf{f}, \tilde{\textbf{u}} ) = Softmax(\textbf{W}_{hy}^ \top \textbf{h}_{t}^{2})
\end{equation}
where $\textbf{W}_{hy}$ $\in \mathbb{R}^{ m \times v}$ , $\textbf{h}_{t}^{2}$  is the hidden state from the second layer of the LSTM network .

\subsubsection{Context Generator}

The clips belong to the same video and are thus contextually co-dependent, \emph{i.e.}, the previous sentences generated by the single sentence generator are semantically relevant to generate the next sentence. To exploit contextual dependencies among the sentences, we build a superordinate context generator that using a LSTM runs over the embedding of each sentence (Figure~3), and is therefore asynchronous to the sentence generator. Each time the state of sentence generator is updated, its output is utilized to initialize the sentence generator of the first layer LSTM:
\begin{equation}
\textbf{x}_{0}^{i} =  {\rm I_{NI}}([{\rm LSTM_c}(\textbf{h}_{M}^{2,i-1},\textbf{h}_{t-1}^{c}),\tilde{\textbf{u}}^{i}] )
\end{equation}
where $\textbf{x}_{0}^{i}$ is the initial input of $i^{th}$ sentence generator. $I_{NI}$ is a dense layer with an activation function, $\textbf{h}_{M}^{2,i-1}$ is the last hidden state of the $i$-$1$$^{th}$ sentence, and $\textbf{h}_{t-1}^{c}$ is the previous hidden state of context LSTM.

In summary, combining visual-language mapping loss $\mathcal{L}(M) $, the cross-entropy loss  $\mathcal{L}(S)$ for description generation can be computed by:
\begin{equation}\label{10}
\mathcal{L}(S) =  - \sum^M_{i=0} \!log(p(\textbf{y}_t| \textbf{y}_{0:t-1})) + \lambda_M \mathcal{L}(M)
\end{equation}
\subsection{Grounding Module}
\noindent$\textbf{Region Grounding}:$ In this section, we aim to evaluate how well the captioning model grounds visual objects. To assist the language model in attending to the correct regions, following ~\cite{zhou2019grounded}, we develop the region attention loss $ \mathcal{L}(R)$: we denote the indicators of positive/negative regions as $\gamma = \{\gamma_1, \cdots, \gamma_N\}$ in each time step, where $\gamma_i =1$ if the region $\gamma_i$ has over 0.5 IoU with a ground truth box and otherwise 0. In combination with the treating attention $a^r$ (Section 3.3.1), the region attention loss function is defined as:
\begin{equation}\label{11}
\mathcal{L}(R)= -\sum _1^N \gamma_i log a_i^r
\end{equation}

\noindent$\textbf{Object Localizaiton}:$ Given an object word $\textbf{w}$ with a specific class label, we aim to localize the related region proposals. We first define the region-class similarity function with the treating attention weights $a^r$ as below:
 \begin{equation}
\begin{aligned}
p^s(\textbf{r}, a^r) &=  {\rm Softmax} (\textbf{W}_s ^ \top \textbf{r}  + a^r )
\end{aligned}
\end{equation}
where $W_s$ $\in \mathbb{R}^{ d \times N}$ is a simple object classifier to estimate the class probability distribution.

Thus we use the $p^s(\textbf{r}, a^r)$ to calculate the confidence score for $\textbf{w}$, combing the supervised attention loss $ \mathcal{L}(R)$, the grounding loss function $\mathcal{L}(G)$ for word $\textbf{w}$ is defined as:
\begin{equation}\label{13}
\mathcal{L}(G)= -\lambda_L\sum ^N_{i=1}\gamma_i log  p^s (\textbf{r}_i) + \lambda_R \mathcal{L}(R)
\end{equation}

\subsection{Training Algorithm}
Algorithm 1 details the pseudocode of our RGL algorithm for GVD generation. First, we pre-train the language GCN by reconstructing the sentences from the latent vector that is encoded from the language $\mathcal{SG}$ with the GCN (sentence $\rightarrow$ $\mathcal{SG}$ $\rightarrow$ latent vector $\rightarrow$ sentence $\rightarrow$ object grounding). Then, we keep the language GCN fixed, and learn the visual GCN by mapping the visual $\mathcal{SG}$  and the language $\mathcal{SG}$  in the latent space. Basically, the encoded latent vector from the language $\mathcal{SG}$  is used as supervised signals to learn the visual GCN and assist the GVD generation. Through the learning procedure, we found that most of the unconcerned visual concepts are filtered.

\renewcommand{\algorithmicrequire}{\textbf{Input:}}  
\renewcommand{\algorithmicensure}{\textbf{Output:}} 
\begin{algorithm}[t]
  \caption{Relational Graph Learning Algorithm}
  \KwIn{Training pairs (${V}$, (${S}$, ${O}^g$ )).}
  \KwOut{ ($  \hat{{S}}, \hat{{O}^g} $).}
   \textbf{Initialization}: Load the pre-trained $\mathcal{SG}$ generator, and generate $\mathcal{SG}^\mathcal{F}$ and
   $\mathcal{SG}^\mathcal{L}$;\\
  \Repeat{Convergence}{ \If{ Language GCN have not trained}
        {
          \Repeat { Convergence;}{
           Randomly sample a minibatch;\\

           ${u}^\mathcal{L}$$\Longleftarrow$ $\mathcal{SG}^\mathcal{L}$ by using Equation (\textcolor{red}{\ref{2}})(\textcolor{red}{\ref{3}})(\textcolor{red}{\ref{4}});\\

           $\bar{{S}}$ $\Longleftarrow$ ${u}^\mathcal{L}$ by using Equation (\textcolor{red}{\ref{10}});\\

           $\bar{{O}^g}$ $\Longleftarrow$  $\bar{{S}}$ by using Equation (\textcolor{red}{\ref{11}}) (\textcolor{red}{\ref{13}});\\

           Update $\Theta$, and minimizing $\mathcal{L}(S)$,$\mathcal{L}(R)$,$\mathcal{L}(G)$;\\
           }

        }
          Fix the Language GCN;\\
          Randomly sample a minibatch;\\
          ${u}^\mathcal{F}$$\Longleftarrow$ $\mathcal{SG}^\mathcal{F}$ by using Equation (\textcolor{red}{\ref{2}})(\textcolor{red}{\ref{3}})(\textcolor{red}{\ref{4}});\\
    Non-linearly map ${u}^\mathcal{F}$ into $\tilde{{u}}$  by using Equation (\textcolor{red}{\ref{5}});\\

           $\hat{{S}}$ $\Longleftarrow$ $\tilde{{u}}$ by using Equation (\textcolor{red}{\ref{10}});\\

           $\hat{{O}^g}$ $\Longleftarrow$  $\hat{{S}}$ by using Equation (\textcolor{red}{\ref{11}}) (\textcolor{red}{\ref{13}});\\
            Update $\Theta$, and minimizing $\mathcal{L}(M)$, $\mathcal{L}(S)$, $\mathcal{L}(R)$, $\mathcal{L}(G)$;
        }

\end{algorithm}
\section{Experiments}
\subsection{Dataset and Setting}
\noindent$\textbf{Dataset}:$ We benchmark the RGL method on the ActivityNet-Entities dataset\footnote{https://github.com/facebookresearch/ActivityNet-Entities} and compare it with other baselines and state-of-the-art models. Moreover, the ActivityNet-Entities collect NPs with bounding box annotations in the frame region level, every NP only in one frame inside each clip. It defines the 4-D tuple $(V, D, E, B)$ that represents the number of videos, descriptions, objects, and bounding boxes, and the official split has \emph{(10k ,35k, 432 ,105k )/,(2.5k, 8.6k, 427, 26.5k)/,( 2.5k ,8.5k, 421, 26.1k)} for train/val/test data, respectively.

\noindent$\textbf{Settings}:$  For video description, we tokenized the texts on white space, and the sentences are ``cut'' at a maximum length of 20 words. All the Arabic numerals are converted to the English word. We add a special \emph{Unknown token} to replace the words out of the vocabulary list. The vocabulary has  4,904 words, and each word represents as a 512-dimensional vector. The hidden state size of the  LSTM is 1024.  The embedding size of nodes in the scene graph and the unified scene graph representation is 1000. The dimension of other learnable matrices are $\textbf{W}_{hy} \in 512 \times 4904$, $\textbf{W}_{hy} \in 2048 \times 1024$. We train our model with cross-entropy objectives and use the ADAM optimizer~\cite{kingma2014adam} with a learning rate of 1.25e-4, the batch size is fixed to 64 when training all models.  All experiments were run on the 2080Ti GPUs. The training is limited to 40 epochs and the model with the best validation CIDEr score is selected for testing.
\subsection{Metrics}
\noindent \textbf{Captioning Evaluation}: We use the performance evaluation tool \footnote{https://github.com/ranjaykrishna/densevid\_eval} provided by the 2018 ActivityNet Captions Challenge, which includes \emph{BLEU@1}, \emph{BLEU@4}, \emph{METEOR}, \emph{CIDEr}, and \emph{SPICE} to evaluate the results of video captioning.
\begin{table*}[t]
  \centering
    \begin{tabular}{l |c| c c c c c| c c c c}
    \hline
     &\multicolumn{1}{|c|}{} &\multicolumn{5}{c|}{ Captioning Evaluation } &\multicolumn{4}{c}{ Grounding Evaluation }\\
    \hline
    Method & RSG. & BLEU@1  & BLEU@4   & METEOR   & CIDEr  & SPICE  & GRD. & ATT.  & F1$_{ALL}$ & F1$_{LOC}$\\
    \hline
    \hline
     GVD (w/o SA)~\cite{zhou2019grounded}$^{\ast}$    && 23.1 &2.16& 10.8& 44.9& 14.9 & 22.3 & 16.1 & 3.73 &11.7\\
    GVD~\cite{zhou2019grounded}$^{\ast}$                  & & 23.6& 2.35& \textbf{11.0}& 45.5& 14.7 & 44.9 & 35.7 & 7.10 & 23.8\\

    \hline
    ST-LSTM$^{\dagger}$         && 22.7 & 2.13 & 10.3 & 42.5 & 11.6 &- &- & - & -\\
    RGL (w/o SG)$^{\dagger}$                &  & 23.4& 2.30& 10.7& 45.9& 14.5     & 45.0 & 35.9 & 7.11 & 23.4\\
     RGL (w/o RA)$^{\dagger}$     &\checkmark  & 25.0& 2.46& 10.7& 45.1& 15.1     & 44.3 & 35.1 &5.29& 14.8\\
    RGL (w/o OG)$^{\dagger}$      &\checkmark  & 25.4& \textbf{2.60}& \textbf{11.1}        & \textbf{47.4} & \textbf{15.4}& \textbf{45.8}& \textbf{36.7} &-& -\\
    RGL (w/o CG)$^{\dagger}$      &\checkmark  & 25.2& \textbf{2.60}& 10.8        & 46.9& \textbf{15.4}& 45.3 & 36.2 &7.69 &26.2\\

     RGL$^{\dagger}$              &\checkmark & \textbf{25.5}& 2.59& 11.0& 47.2& \textbf{15.4}& 45.6 & 36.6 &\textbf{7.70} &\textbf{26.4}\\
    \hline
    \end{tabular}
    \caption{Captioning results and grounding results on ANet-Entities test set.  BLEU-1/4, METEOR, CIDEr and SPICE are used as captioning metrics. Grd., ATT, F1$_{ALL}$ and  F1$_{LOC}$ are used as grounding metrics. RSG. indicates the language refined scene graph is used or not for GVD generation.  $\ast$ indicates the results are obtained from the original papers,  $^{\dagger}$ sentences obtained directly from the author. Larger value indicates better performance. All accuracies are in \%. Top one score on each metric are in bold. Acronym notations of each method see in Sec 4.3.}
\end{table*}

\noindent \textbf{Grounding Evaluation}: Following the grounding evaluation from GVD~\cite{zhou2019grounded} on the generated sentences,  we define the number of object words as $A$, the number of correctly predicted object words as $B$, and the number of correctly predicted and localized words as $C$. A region prediction is considered correct if the object word is correctly predicted and also correctly localized (\emph{i.e.}, IoU with ground truth box \textgreater 0.5). Thus, we compute two versions of precision Fl$_{ALL}$ and Fl$_{LOC}$ to evaluate the object localization accuracy for attention.

   \begin{equation}
\begin{aligned}
 Fl_{ALL}= \frac{C}{A}, \quad \quad Fl_{LOC} = \frac{C}{B}
\end{aligned}
\end{equation}
    During model training, we restrict the grounding region
candidates within the target frame (w/ GT box), i.e., only
consider the N proposals on the frame f with the GT box. We also compute the localization accuracy \emph{GRD.} and attention correctness \emph{ATT.} at each annotated object word.

\noindent \textbf{Hallucination Evaluation}:  Some related object words have not been annotated in the ActivityNet-Entities dataset. Thus, we relabel five related object words for 100 video clips (randomly selected) on the basis of visual content and its ground truth. According to~\cite{rohrbach2018object}, we compute CHAIR$_i$ and CHAIR$_s$, which indicate the fraction of object instances that are hallucinated per-instance and per-sentence, respectively. Besides, we also record the recall of object word prediction  $RECALL_o$.

\noindent \textbf{Human Evaluation}: Since our model can describe a video in more fine-grained details, and some of these details (e.g., Figure~4.a) don't exist in the ground truth. Thus, evaluating the grounding performance of object/relational/attribute words seems unfair. To verify the grounding performance of our model further, we evaluate the description quality through human judgment. On the one hand, the human evaluation  validates the grounded performance of object/relational/attribute words by Relevant$_{obj}$, Relevant$_{rel}$  and Relevant$_{att}$ . On the other hand, we also allow humans to evaluate the coherence of the generated descriptions subjectively using Performance$_{coh}$.

\subsection{Baseline and SoTA} We compare the proposed RGL algorithm with the existing SoTA method GVD~\cite{zhou2019grounded}~\footnote{https://github.com/facebookresearch/grounded-video-description} and baseline ST-LSTM on the ActivityNet-Entities dataset. We also conduct an ablation study to investigate the contributions of individual components in RGL. In our experiment, we train the following baselines of ST-LSTM and four variants of RGL:

  \begin{itemize}

  \item\textbf{ST-LSTM}, which is a captioning model with a two-layer LSTM and spatiotemporal attention and a simplified version of our model.
\item\textbf{RGL (w/o SG)}, which only uses video representation to generate GVD.

\item\textbf{RGL (w/o CG)}, which generates captions without a context generator and merely uses a stack LSTM as its sentence generator.

\item\textbf{RGL (w/o RA)}, which generates captions without regional attention to study the importance of regional supervision.

\item\textbf{RGL (w/o OG)}, which generates captions sequentially without object word grounding and is similar to a standard video captioning algorithm.
 \end{itemize}

\subsection{Experimental Results}

\noindent$\textbf{Captioning Results}:$ Table~1 shows the overall qualitative results of our model and SoTA on the ActivityNet-Entities dataset. Generally, RGL achieves the best performance on all the metrics in comparison with SoTA. We obtain an improvement of 0.7 on SPICE because our method learns the refined representation of $\mathcal{SG}$, which provides relational knowledge and positional semantic prior, to improve this score. It is noteworthy that the RGL (w/o OG) achieves almost all the best captioning scores, this is reasonable, because without grounding operation, the captioning model may pay more attention to the description generation. Moreover, compared with other variants of our method, the performance of RGL (w/o SG) is relatively poor,  thus validating the effectiveness of refined $\mathcal{SG}$  for video description generation.

\noindent$\textbf{Grounding Results}:$ Table~1 also reveals that our method effectively improves the accuracy of GRD., ATT., F1$_{ALL}$ and F1$_{LOC}$. According to our observation, RGL almost obtains the best performance for all of the grounding metrics. Since the $\mathcal{SG}$ representation can be regarded as relational inductive knowledge,  the captioning model can dynamically select relevant information from grounding correct regions and refined $\mathcal{SG}$ to generate correct words.

\noindent$\textbf{Hallucination Results}:$ Table~2 presents object hallucination on the test set.  We note an interesting phenomenon that the $\mathcal{SG}$ based methods tend to perform considerably better on the  CHAIR$_{i}$, CHAIR$_{s}$ and RECALL$_o$ metrics than methods without $\mathcal{SG}$  by a large margin. This finding proves that the $\mathcal{SG}$ contains refined visual concepts that can help the captioning model to capture more correct objects in a video. On this basis, object hallucination is decreased. Therefore, the refined language $\mathcal{SG}$ can assist the region grounding operation in generating correct object words.

\noindent$\textbf{Human Evaluation}:$  As commonly known, the text-matching-based metrics are not perfect, and some descriptions with lower scores actually depict the videos more accurately (See related cases in a later section of Qualitative Examples), i.e., some captioning models can describe a video in detail but with low captioning scores. Thus, human evaluation is conducted. We show five experienced workers the descriptions generated by RGL and RGL (w/o SG) and asked them which one is more coherent and captures more related visual concepts. For each pairwise comparison, a total of 100 video clips (same videos in hallucination evaluation) are randomly extracted for comparison. The results of the comparisons are shown in Table 3. RGL considerably outperforms RGL (w/o SG) in terms of Relevant$_{obj}$, Relevant$_{rel}$  and Relevant$_{att}$ by a large margin. These results indicate that the model with $\mathcal{SG}$  can generate more relevant video description in fine detail.  The results also prove the effectiveness of RGL for GVD generation. Meanwhile, the sentence produced by RGL achieves a higher score in terms of coherent evaluation. This result is obtained probably due to the structured semantic information of $\mathcal{SG}$.

\begin{table}[t]
  \centering
    \begin{tabular}{l |c|c c c  }
    \hline
     & \multicolumn{1}{|c|}{RSG.} & \multicolumn{3}{c}{ Hallucination Evaluation}\\
    \hline
    Method  && CHAIR$_{i}$ & CHAIR$_{s}$ & RECALL$_o$  \\
    \hline
    \hline
     ST-LSTM$^{\dagger}$          & & 0.686 & 0.894& 0.147  \\
    RGL (w/o SG)$^{\dagger}$      && 0.572 & 0.824& 0.198 \\
     RGL (w/o RA)$^{\dagger}$    & \checkmark& 0.659 & 0.867& 0.164 \\
    RGL (w/o OG)$^{\dagger}$     & \checkmark& 0.609 & 0.836& 0.206\\
     RGL (w/o CG)$^{\dagger}$    &\checkmark & 0.531 & 0.807& 0.228 \\
     \hline
      RGL$^{\dagger}$              &\checkmark& \textbf{0.513} & \textbf{0.795}& \textbf{0.234} \\
    \hline
    \end{tabular}
    \caption{ Hallucination results on ANet-Entities test set.  CHAIR$_{i}$,  CHAIR$_{s}$ and ACC$_o$ are used as hallucination metrics. Lower value of CHAIR$_{i}$, CHAIR$_{s}$ indicates better performance, while the large value of ACC$_o$ is better. }
\end{table}

\noindent$\textbf{Analysis of Scene Graph Refinement}:$
To justify the contribution of $\mathcal{SG}$ refinement to GVD generation, we investigate the importance of the $\mathcal{SG}$ with language guidance using three different $\mathcal{SG}$. As we can see in Figure~4, the best performance is obtained when the language $\mathcal{SG}$ is used directly. This result is reasonable because the language  $\mathcal{SG}$ contains key language concepts that can naturally reconstruct its video description.  In addition,   Refined $\mathcal{SG}$ considerably outperforms visual $\mathcal{SG}$  by a large margin on all important metrics, especially the CHAIRi (-9.0). The experimental results indicate the importance of $\mathcal{SG}$ refinement in GVD generation.

\newcommand{\tabincell}[2]{\begin{tabular}{@{}#1@{}}#2\end{tabular}}
\begin{table}[t]
  \centering
    \begin{tabular}{ c c c c c c}
    \hline
    & \multicolumn{3}{c}{ Huamn Evaluation}\\
     \hline
    Metric &      \tabincell{c}{RGL is Better  }   & \tabincell{c}{ RGL is  Worse }  & Equal \\
    \hline
    \hline
    Relevant$_{obj}$ \rule{0pt}{10pt}&0.40 (+9\%) &0.31&0.29  \\
    Relevant$_{rel}$ \rule{0pt}{10pt}&0.43 (+14\%)& 0.29& 0.28 \\
    Relevant$_{att}$ \rule{0pt}{10pt}&0.39 (+16\%) &0.23&0.38  \\
    \hline
    Performance$_{coh}$ \rule{0pt}{10pt}&0.30 (+6\%)&0.24&0.46  \\
    \hline
    \end{tabular}
    \caption{ Human evaluation of our model and its variant RGL (w/o SG).   Relevant$_{obj}$, Relevant$_{rel}$  and Relevant$_{att}$ indicate the relevant visual concepts of objects, relationships and attributes in generated sentence. Performance$_{coh}$ is to evaluate the coherence in sentence. }
\end{table}

\begin{figure}[h]
\includegraphics[width=0.5\textwidth]{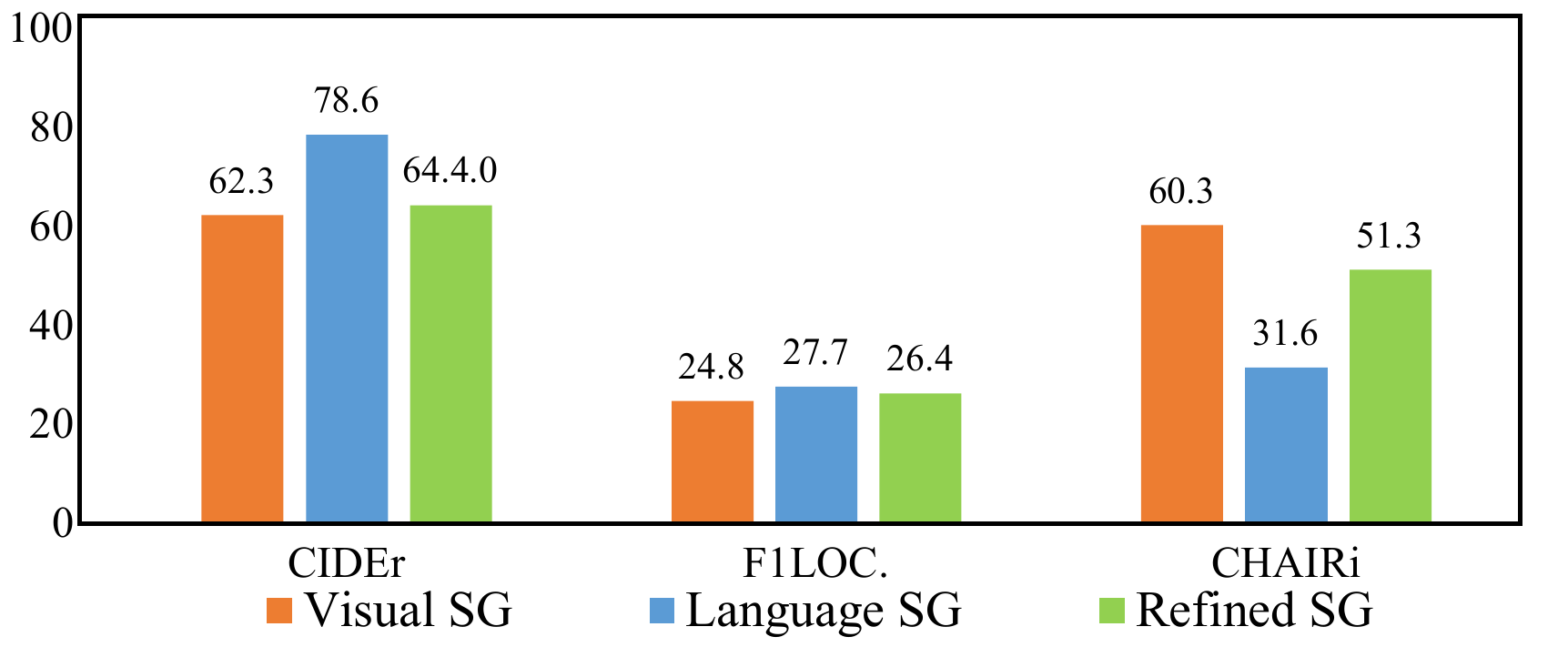}
\centering\caption{The major captioning (CIDEr), grounding (GRD.) and hallucination (CHAIR$_i$) results of using different scene graphs. Visual SG: using video scene graph. Language SG: using the language scene graph. Refined SG: using the language refined scene graph. To facilitate the comparison, the value of CHAIR$_i$ is multiplied by 100.}
\end{figure}
\begin{figure*}[t]
\includegraphics[width=1\textwidth]{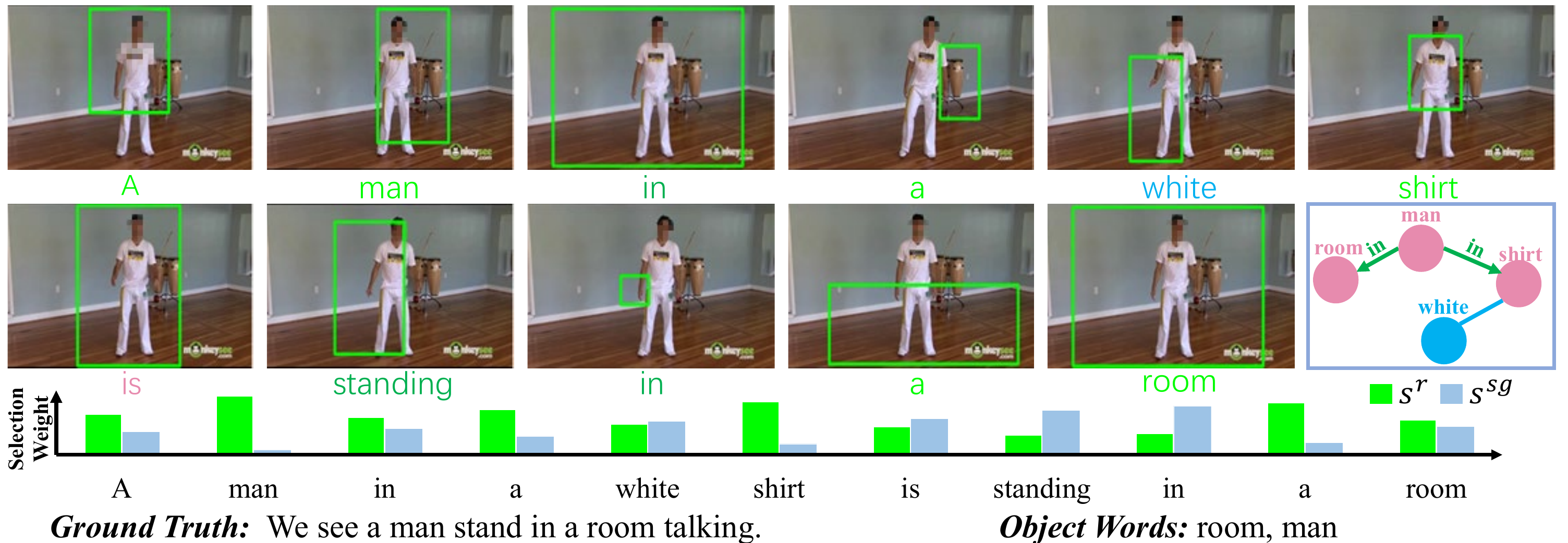}
\vspace{+0.2cm}
\centering\caption{Visualization of words generation. Green texts represent the higher weight of grounding regions. Pink(objects), green(relationships) and blue(attributes) texts indicate that $\mathcal{SG}$ is more important. Green box corresponds to the region with the highest attention $a^r$. Histogram presents the selection weight of $s^{sg}$ and $s^r$.}
\end{figure*}
\begin{figure}[t]
\includegraphics[width=0.5\textwidth]{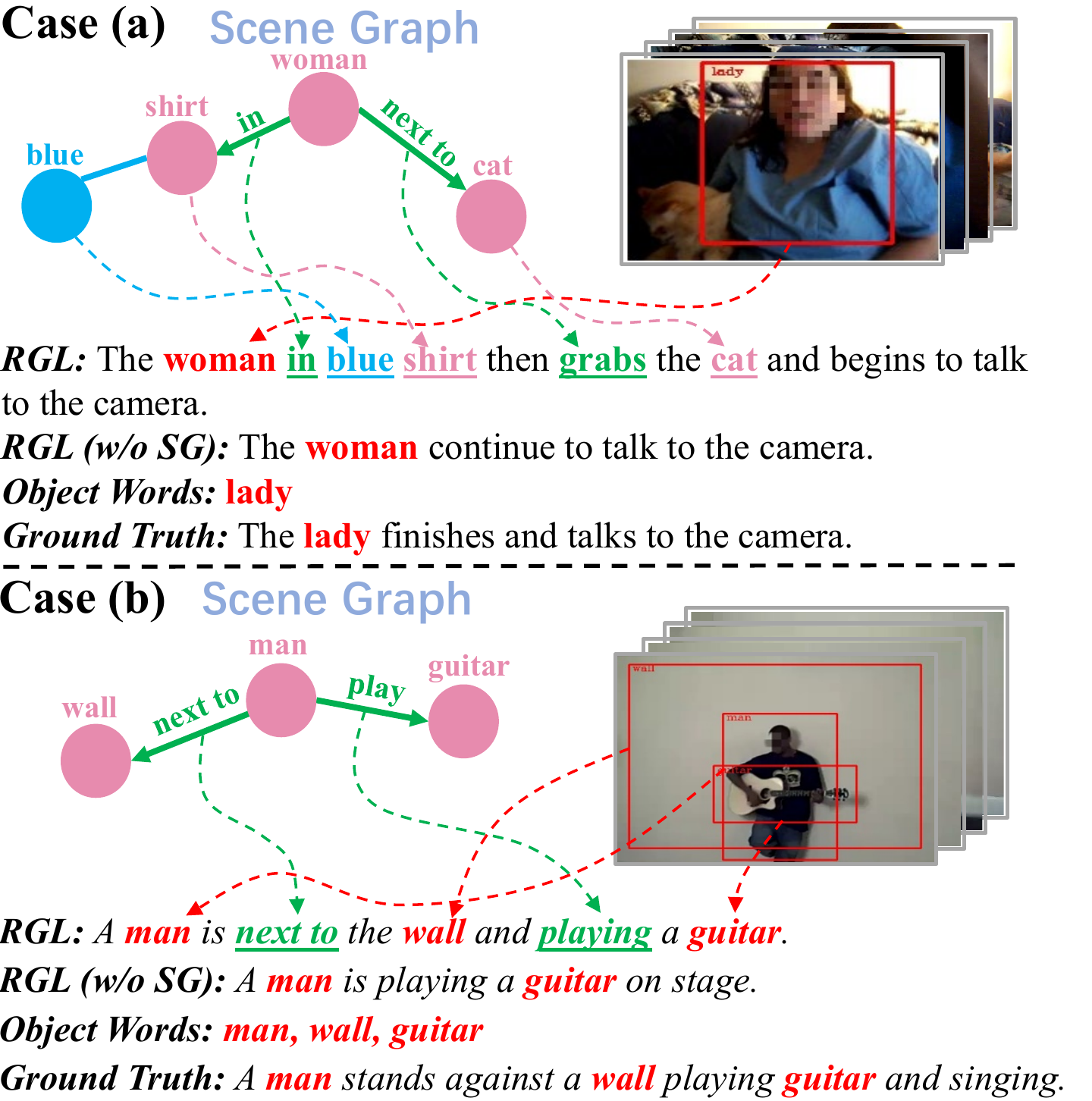}
\centering\caption{Qualitative examples from RGL and RGL (w/o SG). For each figure, the $\mathcal{SG}$ is pruned to avoid clutter. Three word colors correspond to objects, relationships and attributes in detected $\mathcal{SG}$. }

\end{figure}

\noindent$\textbf{Qualitative Examples}:$ To evaluate the quality of video description, we conduct qualitative analysis to compare the RGL and RGL (w/o SG) in Figure~6. We present the $\mathcal{SG}$ for each video, with the annotated bounding boxes as object words. From these exemplary results, RGL produces a more fine-grained caption compared with RGL (w/o SG). As shown in case (a), RGL generates the related objects ``\emph{woman, cat, shirt}",  their relationships ``\emph{in, grab}'', and attribute ``\emph{blue}'' in accordance with the SG, even if the object words only have  ``\emph{lady}''.  Moreover, the relational phrase ``\emph{next to}'' in the $\mathcal{SG}$  is converted to  ``\emph{grab}" through language refinement. However, the sentence generated by RGL (w/o SG) only gives the object word ``\emph{woman}'' to describe the video in a coarse-grained manner, even if it has a higher captioning score and a higher similarity with the ground truth. This result further shows that RGL can generate a more fine-grained and accurate description.

To study how RGL generates a video description, we visualize the selection process during word generation (Figure 5). The histogram presents that the RGL selects \emph{how much} from  $\tilde{r}$ (grounding regions) and  $\tilde{u}$ ($\mathcal{SG}$) to generate each word. The words in green/(pink, green, or blue) color show the RGL select \emph{what} from grounding regions and $\mathcal{SG}$. On the one hand, our model correctly attends to video regions while generating the object words ``\emph{room}'' and ``\emph{man}'', the extra word ``\emph{shirt}'' is also grounded correctly. On the other hand, ``\emph{in}'' and``\emph{white}'', the words in  ``$\langle$\emph{man-in-shirt}$\rangle$ and ``$\langle$\emph{shirt-white}$\rangle$'', the $\mathcal{SG}$ is more important than grounding regions. Thus, the fine-grained phrase ¡°man in white shirt¡± is generated.

\section{Conclusion}
We propose a novel RGL framework to train a GVD generation model. Moreover, we develop a language-refined $\mathcal{SG}$-based method that contains additional visual concepts to describe a video in fine detail. A novel language model with a selection mechanism is also designed. This model can dynamically select the required information from the grounding regions and refined $\mathcal{SG}$ to generate descriptions in a reasonable manner. Our experimental results show the effectiveness of our method via multiple qualitative evaluations. We hope the RGL method can complement the existing literature on video description and benefit further studies on vision and language.

\section{Acknowledgements} 
This work has been supported in part by National Key Research
and Development Program of China (2018AAA010010),
NSFC (U1611461, 61751209, U19B2043, 61976185), Zhejiang
Natural Science Foundation (LR19F020002, LZ17F020001),
2019 Zhejiang University Academic Award for Outstanding
Doctoral Candidates, University-Tongdun Technology Joint
Laboratory of Artificial Intelligence, Zhejiang University
iFLYTEK Joint Research Center, Chinese Knowledge Center
of Engineering Science and Technology (CKCEST), the
Fundamental Research Funds for the Central Universities,
Engineering Research Center of Digital Library, Ministry
of Education. The authors from UCSC and UCSB are not
supported by any of the projects above.
\bibliographystyle{ACM-Reference-Format}
\bibliography{ref}


\end{document}